\RequirePackage[hyphens]{url}
\documentclass[11pt,a4paper]{article}
\usepackage[hyperref]{acl2021}
\usepackage{times}
\usepackage{latexsym}

\usepackage{microtype}

\usepackage{hyperref}
\PassOptionsToPackage{hyphens}{url}\usepackage{hyperref}
\usepackage{float} 
\usepackage{boldline,multirow}
\usepackage{makecell}
\usepackage{amsmath}
\usepackage{multicol}
\usepackage{booktabs}
\usepackage{graphicx}
\usepackage{csquotes}
\usepackage{xcolor}
\usepackage{xspace}
\aclfinalcopy 

\newcommand{\ie}{i.e.\@\xspace}
\newcommand{\eg}{e.g.\@\xspace}

\newcommand{\engl}[1]{\textcolor{gray}{\small [#1]}}

\title{Modeling Profanity and Hate Speech in Social Media\\with Semantic Subspaces}

\author{Vanessa Hahn, Dana Ruiter, Thomas Kleinbauer, Dietrich Klakow \\
  Spoken Language Systems Group\\
  Saarland University\\
  Saarbr\"{u}cken, Germany\\
  \texttt{\{vhahn|druiter|kleiba|dklakow\}@lsv.uni-saarland.de}}

\begin{document}
\maketitle
\begin{abstract}
Hate speech and profanity detection suffer from data sparsity, especially for languages other than English, due to the subjective nature of the tasks and the resulting annotation incompatibility of existing corpora.  In this study, we identify profane subspaces in word and sentence representations and explore their generalization capability on a variety of similar and distant target tasks in a zero-shot setting. This is done monolingually (German) and cross-lingually to closely-related (English), distantly-related (French) and non-related (Arabic) tasks. We observe that, on both similar and distant target tasks and across all languages, the subspace-based representations
transfer more effectively than standard BERT representations
in the zero-shot setting, with 
improvements
between F1 $+10.9$ and F1 $+42.9$ 
over the baselines
across all tested monolingual and cross-lingual scenarios.
\end{abstract}

\section{Introduction}
Profanity and online hate speech have been recognized as crucial problems on social media platforms as they bear the potential to offend readers and disturb communities. The large volume of user-generated content makes manual moderation very difficult and has motivated a wide range of natural language processing (NLP) research in recent years. However, the issues are far from solved, and the automatic detection of profane and hateful contents in particular faces a number of severe challenges.

Pre-trained transformer-based \citep{vaswani2017attention} language models, \eg BERT \citep{devlin-etal-2019-bert}, play a dominant role today in many NLP tasks. However, they work best when large amounts of training data are available. This is typically not the case for profanity and hate speech detection where few datasets are currently available \citep{waseem-hovy:2016:N16-2,basile-etal-2019-semeval,Struss2019GermEval} with moderate sizes at most. In addition, these tasks are known to be highly subjective \citep{waseem:2016:NLPandCSS}. Annotation protocols for hate speech and profanity often rely on different assumptions that make it non-trivial to combine multiple datasets. In addition, such datasets only exist for few languages besides English \citep{ousidhoum-etal-2019-multilingual,abu-farha-magdy-2020-multitask,zampieri-etal-2020-semeval}.

For such low-resource scenarios, few- and zero-shot transfer learning has seen an increased interest in the research community. One particular approach, using semantic subspaces to model specific linguistic aspects of interest \citep{rothe-etal-2016-ultradense}, has proven to be effective for representing contrasting semantic aspects of language such as \eg positive and negative sentiment.

In this paper, we propose to learn \textbf{semantic subspaces to model profane language} on both the word and the sentence level. This approach is especially promising because of its ability to cope with
sparse profanity-related datasets confined to very few languages. Profanity and hate speech often co-occur but are not equivalent, since not all hate speech is profane (e.g. \emph{implicit} hate speech) and not all profanity is hateful (e.g. \emph{colloquialisms}). Despite being \emph{distantly related} tasks, we posit that modeling profane language via semantic subspaces may have a positive impact on downstream hate speech tasks.

We analyze the efficacy of the subspaces to encode the profanity (\emph{neutral} vs. \emph{profane} language) aspect and apply the resulting subspace-based representations to a \textbf{zero-shot transfer classification} scenario with both similar (\emph{neutral}/\emph{profane}) and distant (\emph{neutral}/\emph{hate}) target classification tasks. To study their ability to generalize across languages we evaluate the zero-shot transfer in both a \textbf{monolingual} (German) and a \textbf{cross-lingual} setting with closely-related\footnote{Both English and German belong to the West-Germanic language branch, and are thus closely-related. French, on the other hand, is only distantly related to German via the Indo-European language family, while Arabic (Semitic language family) and German are not related.} (English), distantly-related (French) and non-related (Arabic) languages.

We find that subspace-based representations outperform popular alternatives, such as BERT or word embeddings, by a large margin across all tested transfer tasks, indicating their strong generalization capabilities not only monolingually but also cross-lingually.
We further show that semantic subspaces can be used for \textbf{word-substitution} tasks 
with the goal of generating automatic suggestions of neutral counterparts for the civil rephrasing of profane contents.

\section{Related Work}
\label{s:related_work}

\textbf{Semantic subspaces} have been used to identify gender \citep{bolukbasi2016debiasing} or multiclass ethnic and religious \citep{manzini-etal-2019-black} bias in word representations. \citet{liang2019sentdebias} identify multiclass (gender, religious) bias in sentence representations. Similarly, \citet{carpuat2017style} identify a stylistic subspace that captures the degree of formality in a word representation. This is done using a list of minimal-pairs, \ie pairs of words or sentences that only differ in the semantic feature of interest over which they perform principal component analysis (PCA). We take the same general approach in this paper (see Section \ref{s:method}). 

Conversely, \citet{gonen-goldberg-2019-lipstick-pig} show that the methods in \citet{bolukbasi2016debiasing} are not able to identify and remove the gender bias entirely. Following this, \citet{ravfogel-etal-2020-null} argue that semantic features such as gender are encoded non-linearly, and suggest an iterative approach to identifying and removing gender features from semantic representations entirely.

Addressing the issue of data sparseness,
\citet{rothe-etal-2016-ultradense} use ultradense subspaces to generate task-specific representations that capture semantic features such as abstractness and sentiment and show that these are especially useful for low-resourced downstream tasks. 
While they focus on using small amounts of labeled data of a specific target task to learn the subspaces, we focus our study on learning a generic profane subspace and test its generalization capacity on similar and distant target tasks in a zero-shot setting.

\textbf{Zero-shot transfer}, where a model trained on a set of tasks is evaluated on a previously unseen task, has recently gained a lot of traction in NLP. 
Nowadays, this is done using large-scale transformer-based language models such as BERT, that share parameters between tasks. 
Multilingual varieties such as XLM-R \citep{conneau-etal-2020-unsupervised} 
enable the zero-shot cross-lingual transfer of a task. One example is sentence classification trained on a (high-resource) language being transferred into another (low-resource) language \citep{hu2020xtreme}.

\section{Method: Semantic Subspaces}
\label{s:method}

A common way to represent word-level semantic subspaces is based on a set $P$ of so-called \emph{minimal pairs}, \ie 
$N$ 
pairs of words 
$(w,\hat{w})$
that differ only in the semantic dimension of interest \cite{bolukbasi2016debiasing,carpuat2017style}. 
Table~\ref{t:word_class} displays some examples of such word pairs for the profanity domain.
\begin{table}
\centering
\begin{tabular}{ll}
\toprule
$\mathbf{w}$ \textbf{(profane)} & 
$\mathbf{\hat{w}}$ \textbf{(neutral)} \\ \midrule
Arschloch \engl{asshole} & Mann \engl{man} \\
Fotze \engl{cunt} & Frau \engl{woman} \\
Hackfresse \engl{shitface} & Mensch \engl{human} \\
\bottomrule
\end{tabular}\caption{
Examples of word-level minimal pairs.}
\label{t:word_class}
\vspace{-4mm}
\end{table}
Each word $w$ is encoded as a word embedding $e(w)$: 
\begin{equation*}
P = \{ (e(w_1), e(\hat{w}_1)), \dots, (e(w_N), e(\hat{w}_N)) \}
\end{equation*}
Then, each pair is normalized by a mean-shift:
\begin{equation*}
\bar{P} = \{ (e(w_i)-\mu_i, e(\hat{w}_i)-\mu_i) | 1 \leq i \leq N \}
\end{equation*}
where each $\mu_i = \frac{1}{2} (e(w_i)+e(\hat{w}_i))$.

Finally, PCA is performed on the set $\bar{P}$ and the most significant principal component (PC) is used as a representation of the semantic subspace.

\medskip\noindent
We diverge from this approach in four ways:

\paragraph{Normalization}
We note that there is no convincing justification for the normalization step. As our experiments in the following sections show, we find that the profanity subspace is better represented by $P$ than by $\bar{P}$.
For our experiments, we thus distinguish three different types of representations:

\begin{itemize}
    \item \textbf{BASE}: The raw featurized representation $r$.
    \item \textbf{PCA-RAW}: Featurized representation $r$ projected onto the non-normalized subspace $S(P)$.
    \item \textbf{PCA-NORM}: Featurized representation $r$ projected onto the normalized subspace $S(\bar{P})$.
\end{itemize}

Here, projecting a vector representation $r$ onto a subspace is defined as the dot product $r \cdot S(P)$.

\paragraph{Number of Principal Components $\mathbf{c}$}
The use of just a single PC as the best representation of the semantic subspace is not well motivated. This is recognized by \citet{carpuat2017style} who experiment on the first $c=1, 2, 4,\dots,512$ PC and report results on their downstream-task directly. However, a downside of their method for determining a good value for $c$ is the requirement of a task-specific validation set which runs orthogonal to the assumption that a good semantic subspace should generalize well to many related tasks.

Instead, we propose the use of an \emph{intrinsic evaluation} that requires no additional data 
to estimate a good value for $c$. \citet{rothe-etal-2016-ultradense} have shown that semantic subspaces are especially useful for classification tasks related to the semantic feature encoded in the subspace. 
Here, we argue the inverse: if a semantic subspace with $c$ components yields the best performance on a related classification task, $c$ should be an appropriate number of components to encode the semantic feature. 

More specifically, we apply a classifier function $f(x)=y$, which learns to map a subspace-based representation $x = e \cdot S(P)$ to a label $y \in \{{\rm profane, neutral}\}$. We learn $f(x)$ on the same set $P$ used to learn the subspace. In order to evaluate on previously unseen entities, we employ 5-fold cross validation over the available list of minimal pairs $P$ and evaluate Macro F1 on the held-out fold. Due to the simplicity of this intrinsic evaluation, the experiment can be performed for all values of $c$ and the $c$ yielding the highest average Macro F1 is selected as the final value. The above holds for $P$ and $\bar{P}$ equally.

\paragraph{Sentence-Level Minimal Pairs}

We move the word-level approach to the sentence level. In this case, minimal pairs are made up of vector representations of sentences $(e(s), e(\hat{s}))$.

In order to standardize the approach and to focus the variation in the sentence representations on the profanity feature, sentence-level minimal pairs are constructed by keeping all words contained equivalent except for \emph{significant words} that in themselves are minimal pairs for the semantic feature of interest. For instance, a sentence-level minimal pair for the \emph{profanity} feature with \underline{significant} words:

\begin{displayquote}
\textit{The food here is \underline{shitty}. \\
The food here is \underline{disgusting}.}
\end{displayquote}

\paragraph{Zero-Shot Transfer}

In order to evaluate how well profanity is encoded in the resulting word- and sentence-level subspaces, we test their generalization capabilities in a zero-shot classification setup. Given a subspace $S(P)$ (or $S(\bar{P})$), we train a classifier $f(x)=y$ to classify subspace-based representations $x = e \cdot S(P)$ as belonging to class $y \in \{{\rm profane|neutral}\}$. The $x$ used to train the classifier are the same entities in the minimal pairs used to learn $S(P)$.
This classification task is the \emph{source task} $\mathcal{T} = \{x,y\}$. As the classifier is learned on subspace-based representations, it should be able to generalize significantly better to previously unseen profanity-related tasks than a classifier learned on generic representations $x = e$ \citep{rothe-etal-2016-ultradense}. 
Given a previously unseen task $\mathcal{\bar{T}} = \{\bar{x},\bar{y}\}$, we follow a \textbf{zero-shot transfer} approach and let classifier $f$, learned on source task $\mathcal{T}$ only, predict the new labels $\bar{y}$ given instances $\bar{x}$ without training it on data from $\mathcal{\bar{T}}$. The zero-shot generalization can be quantified by calculating the accuracy of the predicted labels $\hat{\bar{y}}$ given the gold labels $\bar{y}$.
The extend of this zero-shot generalization capability can be tested by performing zero-shot classification on a variety of unseen tasks $\mathcal{\bar{T}}$ with variable task distances $\mathcal{\bar{T}} \Leftrightarrow \mathcal{T}$.

\section{Experimental Setup}
\label{s:experimental_setup}

\subsection{Data}
\label{s:data}

\paragraph{Word Lists} The minimal-pairs used in our experiments are derived from a German slur collection\footnote{\url{www.hyperhero.com/de/insults.htm}}.

\paragraph{Fine-Tuning} We use the German, English, French and Arabic portions of a large collection of tweets\footnote{\url{www.archive.org/details/twitterstream}} collected between 2013--2018 to fine-tune BERT. For the German BERT model, all available German tweets are used, while the multilingual BERT is fine-tuned on a balanced corpus of 5M tweets per language. For validation during fine-tuning, we set aside 1$k$ tweets per language.

\begin{table}[t]
\small
\centering
\begin{tabular}{l r r}
\toprule
\textbf{Corpus} & \textbf{\# Sentences} & \textbf{\# Tokens} \\ 
      \midrule
\multicolumn{2}{l}{\emph{Fine-Tuning}} \\
Twitter-DE & 5(9)M & 45(85)M\\
Twitter-EN &  5M & 44M \\
Twitter-FR & 5M & 58M \\
Twitter-AR &  5M & 75M \\ \midrule
\multicolumn{2}{l}{\emph{Target Tasks}} \\
DE-ST & 111/111 & 1509/1404 \\
DE-DT & 2061/970 & 14187/9333 \\
EN-ST & 93/93 & 1409/1313 \\
EN-DT & 288/865  & 8032/3647 \\
AR-ST & 12/12  & 164/84 \\
AR-DT & 46/54 & 592/506 \\
FR-DT & 5822/302 & 49654/2660 \\
\bottomrule
\end{tabular}
\caption{Number of sentences and tokens of the data used for fine-tuning BERT for the sentence-level experiments. Target task test sets are reported with their respective \textit{neutral}/\textit{hate} (DT) and \textit{neutral}/\textit{profane} (ST) distributions. 
}
\label{t:corpora}
\end{table}

\paragraph{Target Tasks} We test our sentence-level representations, which are used to train a \textit{neutral}/\textit{profane} classifier on a subset of minimal pairs, on several hate speech benchmarks. For all four languages, we focus on a distant task DT (\textit{neutral}/\textit{hate}). For German, English and Arabic we additionally evaluate on a similar task ST (\textit{neutral}/\textit{profane}), for which we removed additional classes (\textit{insult}, \textit{abuse} etc.) from the original finer-grained data labels and downsampled to the minority class (\textit{profane}).

For German (DE), we use the test sets of GermEval-2019 \citep{Struss2019GermEval} Subtask 1 (\textit{Other}/\textit{Offense}) and Subtask 2 (\textit{Other}/\textit{Profanity}) for DT and ST respectively. For English (EN), we use the HASOC \citep{mandl2019hasoc} Subtask A (\textit{NOT}/\textit{HOF}) and Subtask B (\textit{NOT}/\textit{PRFN}) for DT and ST respectively. French (FR) is tested on the hate speech portion (\textit{None}/\textit{Hate}) of the corpus created by \citet{charitidis2019towards} for DT only, while Arabic (AR) is tested on \citet{mubarak-etal-2017-abusive} for DT (\textit{Clean}/\textit{Obscene+Offense}) and ST (\textit{Clean}/\textit{Obscene}). As AR has no official train/test splits, we use the last 100 samples for testing. The training data of these corpora is not used.

Table \ref{t:corpora} summarizes the data used for fine-tuning as well as testing.

\paragraph{Pre-processing}
The Twitter corpora for fine-tuning were pre-processed by filtering out incompletely loaded tweets and duplicates. We also applied language detection using \texttt{spacy} to further remove tweets that consisted of mainly emojis or tweets that were written in other languages. 

\subsection{Model Specifications}
\label{s:model_specifications}

To achieve good coverage of profane language, we use 300-dimensional German FastText embeddings \citep{deriu2017mlsent} trained on 50M German tweets for the word-level experiments in Section \ref{s:word_level}.

The BERT models \citep{devlin-etal-2019-bert} used in Section \ref{s:sentence_level} are \texttt{Bert-Base-German-Cased}\footnote{\url{www.deepset.ai/german-bert}} and \texttt{Bert-Base-Multilingual-Cased} for the monolingual and multilingual experiments respectively, since they pose strong baselines. 
We fine-tune on the Twitter data (Section \ref{s:data}) using the masked language modeling objective
and
early stopping over the evaluation loss ($\delta=0$, ${\rm patience}=3$). 
All classification experiments use Linear Discriminant Analysis (LDA) as the classifier.

\section{Word-Level Subspaces}
\label{s:word_level}

Before moving to the lesser explored sentence-level subspaces, we first verify whether word-level semantic subspaces can also capture complex semantic features such as profanity.

\subsection{Minimal Pairs}
\label{s:word_minimal_pairs}

\begin{figure*}[t]
    \centering
    \includegraphics[trim=0 50 0 0,clip,width=\textwidth]{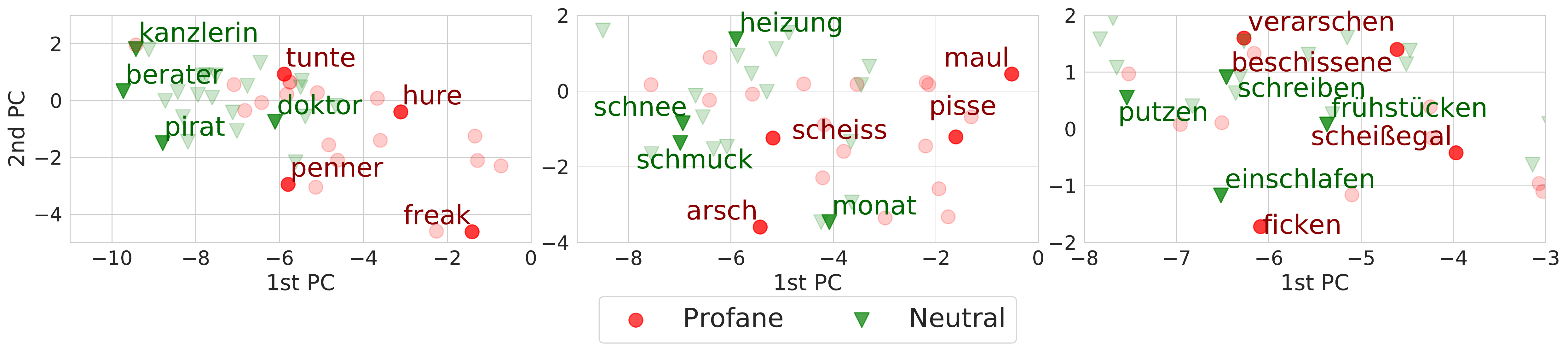}\\
    \caption{Projections of profane and neutral words from TL-1 (left), TL-2 (middle) and TL-3 (right) onto a word-level profane subspace learned by PCA-NORM on 10 minimal pairs (%
      \includegraphics[trim=570 20 844 280,clip,scale=0.3165]{projections4.pdf} Profane,
      \includegraphics[trim=780 20 638 280,clip,scale=0.3165]{projections4.pdf} Neutral).
    }
    
    \label{f:word_projection}
\end{figure*}

Staying within the general low-resource setting prevalent in hate speech and profanity domains, 
and to keep manual annotation effort low, we randomly sample a small amount of words from the German slur lists, namely 100, and manually map these to their neutral counterparts (Table \ref{t:word_class}). We focus this list on nouns describing humans.

Each word in our minimal pairs is featurized using its word embedding, this is our BASE representation. We learn PCA-RAW and PCA-NORM representations on the embedded minimal pairs.

\subsection{Classification}
\label{s:classification}

We evaluate how well the resulting representations BASE, PCA-RAW and PCA-NORM encode information about the profanity of a word by focusing on a related word classification task where unseen words are classified as \textit{neutral} or \textit{profane}. To evaluate how efficient the subspaces can be learned in a low-resource setting, we downsample the list of minimal pairs to learn the subspace-based representations and the classification task to 10--100 word pairs. 
After the preliminary exploration of the number of principal components (PC) required to represent profanity, the number of PC for the final representations lie within a range of 15--111. 
Each experiment is run over 5 seeded runs and we report the average F1 Macro with standard error. As each seeded run resamples the training and test data, the standard error is also a good indicator of the variability of the method when trained on different subsets of minimal pairs.

\paragraph{Test Lists}
For this evaluation, we create three test lists (TL-\{1,2,3\}) of profane and neutral words.
The contents of the three TLs are defined by their decreasing relatedness to the list of minimal pairs used for learning the subspace, which are nouns describing humans. 
TL-1 is thus also a list of nouns describing humans, TL-2 contains random nouns not describing humans, and TL-3 contains verbs and adjectives. 
The three TLs are created by randomly sampling from the word embeddings that underlie the subspace representations and adding matching words to TL-\{1,2,3\} until they each contain 25 profane and 25 neutral words, \ie 150 in total.

Projecting the TLs onto the first and second PC of the PCA-NORM subspace learned on 10 minimal pairs suggests that a separation of profane and neutral words can be achieved for nouns describing humans (TL-1), while it is more difficult for less related words (TL-\{2,3\}) (Figure \ref{f:word_projection}).

\begin{figure*}[t]
    \centering
    \includegraphics[trim=0 44 0 0,clip,width=\textwidth]{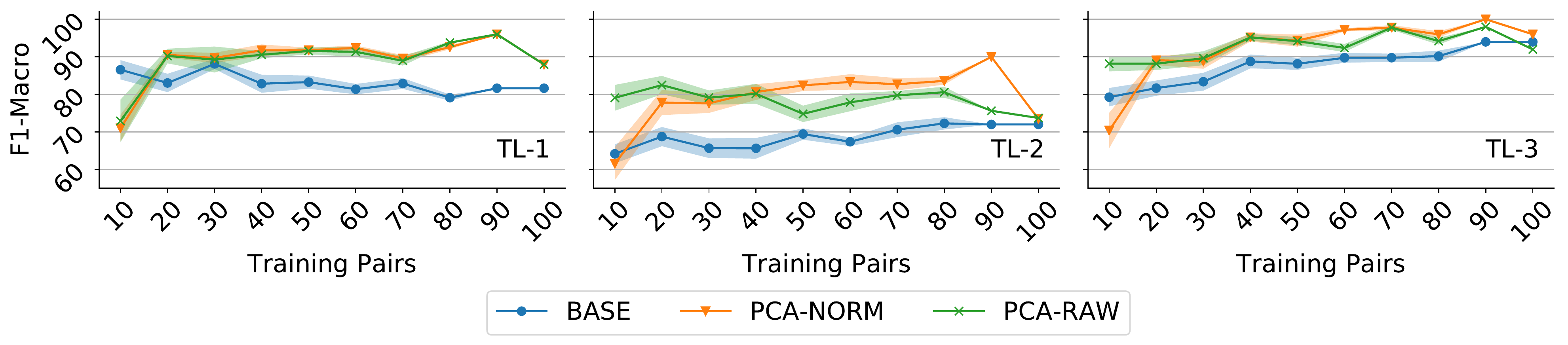}\\
    \caption{F1-Macro of the LDA models, using BASE or PCA-\{RAW,NORM\} representations on the word classification task based on 10 to 100 training word pairs (
    \includegraphics[trim=360 18 740 218,clip,scale=0.4]{graph3.pdf} BASE,
    \includegraphics[trim=490 18 606 218,clip,scale=0.4]{graph3.pdf} PCA-NORM,
    \includegraphics[trim=678 18 424 218,clip,scale=0.4]{graph3.pdf} PCA-RAW).    
    }
    \label{f:wordclassification_results}
    \vspace{-2mm}
\end{figure*}

\paragraph{Results}
Across all TLs, the subspace-based representations outperform the generalist BASE representations (Figure \ref{f:wordclassification_results}), with PCA-NORM reaching F1-Macro scores of up to $96.0$ (TL-1), $89.9$ (TL-2) and $100$ (TL-3) when trained on 90 word pairs. This suggests that they generalize well to unseen nouns describing humans as well as verbs and adjectives, while generalizing less to nouns not describing humans (TL-2). This may be due to TL-2 consisting of some less frequent compounds (\eg Gro{\ss}maul \engl{big mouth}). PCA-NORM and PCA-RAW perform equally on TL-1 and TL-3, while PCA-NORM is slightly stronger on the mid-resource (50-90 pairs) range on TL-2. This suggests that the normalization step when constructing the profane subspace is only marginally beneficial. Even when the training data is very limited (10--40 pairs), the standard errors are decently small (F1 $\pm1$--$5$), indicating that the choice of minimal pairs has only a small impact on the downstream model performance. When more training data is available (80--100 pairs), the influence of a single minimal pair becomes less pronounced and thus the standard error decreases significantly.

\subsection{Substitution}
\label{s:word_substitution}
\begin{table*}
\centering
\small
\renewcommand{\arraystretch}{1.1}
\begin{tabular}{lll}
  \toprule
  
  \textbf{Word $w$} & \textbf{NN($w$)} & \textbf{NN($\hat{w}$)} \\ 
  
  \midrule
  
  \makecell[tl]{Scheisse \\ \engl{shit}} &
  Scheiße, Scheissse, Scheissse04, Scheißee &
  \makecell[tl]{schrecklich, augenscheinlich, schwerlich, schwesterlich \\
               \engl{horrible, evidently, hardly, sisterly}} \\

  \makecell[tl]{Spast \\ \engl{dumbass}} &
  \makecell[tl]{Kackspsst, Spasti, Vollspast, Dummerspast} &
  \makecell[tl]{Mann, Mensch, Familienmensch, Menschn \\
               \engl{man, person, family person, people}} \\

  Bitch & 
  \makecell[tl]{x6bitch, bitchs, bitchin, bitchhh} & 
  \makecell[tl]{Frau, Afrikanerin, Mann, Amerikanerin \\ 
                \engl{woman, african, man, american}} \\

  \makecell[tl]{Arschloch \\ \engl{asshole}} & 
  \makecell[tl]{Narschloch, Arschlochs, Arschloc, learschloch} & 
  \makecell[tl]{Mann, Frau, Lebenspartnerin, Menschwesen\\
                \engl{man, woman, significant other, human creature}} \\

  \makecell[tl]{Fresse \\ \engl{cakehole}} & 
  \makecell[tl]{Fresser, Schnauze, Kackfufresse, Schnauzefresse} & 
  \makecell[tl]{Frau, Mann, Lebensgefährtin, Rentnerin\\
                \engl{woman, man, significant other, retiree}} \\
  \bottomrule
\end{tabular}

\caption{Profane words $w$ with top 4 NNs before (${\rm NN}(w)$) and after (${\rm NN}(\hat{w})$) removal of the profane subspace.}
\label{t:word_nn}
\end{table*}

We use the profane subspace $S_{\rm prf}$ to substitute a profane word $w$ with a neutral counterpart $\hat{w}$. We do this by removing $S_{\rm prf}$ from $w$,

\begin{align}
    \hat{w} &= \frac{w - S_{\rm prf}}{||w - S_{\rm prf}||}
\end{align}

and replacing it by its new nearest neighbor ${\rm NN}(\hat{w})$ in the word embeddings. Here, we focus on the PCA-NORM subspace learned on 10 minimal pairs only. We use this subspace to substitute all profane words in TL-\{1,2,3\}.

\paragraph{Human Evaluation}
To analyze the similarity and profanity of the substitutions, we perform a small human evaluation.
Four annotators were asked to rate the similarity of profane words and their substitutions, and also to give a profanity score between 1 (not similar/profane) and 10 (very similar/profane) to words from a mixed list of slurs and substitutions.

Original profane words were rated with an average of 6.1 on the \textbf{profanity} scale, while substitutions were rated significantly lower, with an average rating of 1.9. Minor differences exist across TL splits, with TL-1 dropping from 6.8 to 1.3, TL-2 from 6.1 to 3.1 and TL-3 from 5.4 to 2.1.

The average \textbf{similarity} rating between profane words and their substitution differs strongly across different TLs. TL-1 has the lowest average rating of 2.8, while TL-2 has a rating of 3.3 and TL-3 a rating of 5.1. This is surprising, since the subspaces generalized well to TL-1 on the classification task.

\paragraph{Qualitative Analysis}
To understand the quality of the substitutions, especially on TL-1, which has obtained the lowest similarity score in the human evaluation, we perform a small qualitative analysis on 3 words sampled from TL-1 (\emph{Spast, Bitch, Arschloch}) and 1 word sampled from TL-2 (\emph{Fresse}) and TL-3 (\emph{Scheiss}) each.
Before removal, the nearest neighbors (NNs, Table \ref{t:word_nn}) of the sampled offensive words were mostly orthographic variations (\eg \textit{Scheisse \engl{shit}} vs. \textit{Scheiße}) or compounds of the same word (\eg \textit{Spast \engl{dumbass}} vs. \textit{Vollspast \engl{complete dumbass}}). After removal, the NNs are still negative but not profane (\eg \textit{Scheisse} \textrightarrow \textit{schrecklich \engl{horrible}}). While the first NNs are decent counterparts, later NNs introduce other (gender, ethnic, etc.) biases,
possibly stemming from the word embeddings or from the minimal pairs used to learn the subspace.
The counterparts to \emph{Scheisse \engl{shit}} seem to focus around the phonetics of the word (all words contain \emph{sch}), which may also be due to the poor representation of adjectives in embedding spaces. \emph{Fresse \engl{cakehole}} is ambiguous\footnote{\emph{Fresse} can mean \emph{shut up}, as well as being a pejorative for \emph{face} and \emph{eating}.}, thus the subspace does not entirely capture it and the new NNs are neutral, but unrelated words.

While human similarity ratings on TL-1 were low, qualitative analysis shows that these can still be reasonable. The low rating on TL-1 may be due to annotators' reluctance to equate human-referencing slurs to neutral counterparts.

The ability to automatically find neutral alternatives to slurs may lead to practical applications such as the suggestion of alternative wordings.

\section{Sentence-Level Subspaces}
\label{s:sentence_level}

In Section \ref{s:word_level}, we identified profane subspaces on the word-level. 
However, abuse mostly happens on the sentence and discourse-level and is not limited to the use of isolated profane words.
Therefore, we move this method to the sentence-level, exploring the two subspace-based representation types PCA-RAW and PCA-NORM. 
Concretely, we learn sentence-level profane subspaces that allow a context-sensitive representation and thus go beyond isolated profane words, and verify their efficacy to represent \emph{profanity}. Similarly to the word-level experiments, we focus our analysis on the ability of the subspaces to generalize to similar (\textit{neutral}/\textit{profane}) and distant (\textit{neutral}/\textit{hate}) tasks. We compare their performance with a BERT-encoded BASE representation, which does not use a semantic subspace.

\subsection{Minimal Pairs}

Using the German slur collection, we identify tweets in Twitter-DE containing swearwords, from which we then take $100$ random samples. 
We create a neutral counterpart by manually replacing significant words, \ie swearwords, with a neutral variation while keeping the rest of the tweet as is: 

\begin{displayquote}
\begin{tabular}{l@{\hspace{2mm}}l}
a) & \textit{ich darf das nicht verkacken!!!}\\
   & \textit{\engl{I must not fuck this up!!!]}} \\
b) & \textit{ich darf das nicht vermasseln!!!}\\
   & \textit{\engl{I must not mess this up!!!}}\\
\end{tabular}
\end{displayquote}

\subsection{Monolingual Zero-Shot Transfer}
\label{s:monolingual}

We validate the generalization of the German sentence-level subspaces to a similar (\textit{profane}) and distant (\textit{hate}) domain by zero-shot transferring them to unseen German target tasks and analyzing their performance.

\subsubsection{Representation Types}
\label{s:monolingual_representation}

We fine-tune \texttt{Bert-Base-German-Cased} on Twitter-DE (9M Tweets). Each sentence in our list of minimal pairs is then encoded using the fine-tuned German BERT and its sentence representation $s = {\rm mean}(\{h_1,...,h_T\})$ is the mean over the $T$ encoder hidden states $h$.
This is our BASE representation. We further train PCA-RAW and PCA-NORM on a subset of our minimal pairs. 
We chose 14--96 PCs for PCA-RAW and 9--94 PCs for PCA-NORM depending on the size of the subset of minimal pairs used to generate the subspace. 

\subsubsection{Results}
\label{s:monolingual_results}

We train the PCA-RAW and PCA-NORM representations on subsets of increasing size ($10, 20, \dots, 100$ minimal pairs). For each subset and representation type (BASE, PCA-RAW, PCA-NORM), we train an LDA model to identify whether a sentence in the subset of minimal pairs is neutral or profane.
These models are zero-shot transferred to the German similar task ST (\textit{neutral}/\textit{profane}) and distant task DT (\textit{neutral}/\textit{hate}). We report the average F1-Macro and standard error over $5$ seeded runs, where each run resamples its train and test data.

\paragraph{ST: Similar Task}

\begin{figure}[t]
    \centering
    \includegraphics[trim=0 60 0 0,clip,width=\columnwidth]{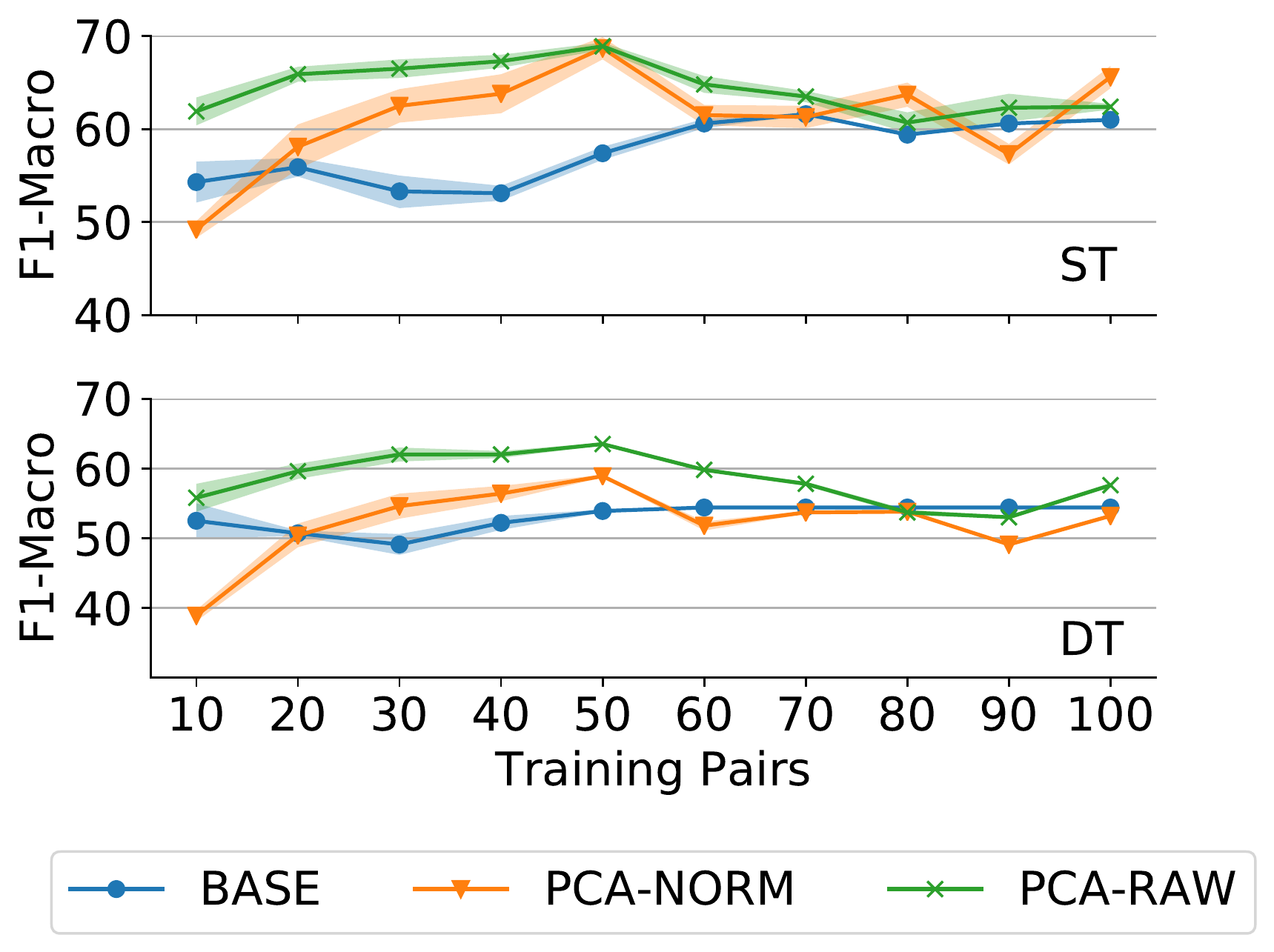}
    \caption{F1-Macro of the LDA models, 
      zero-shot transferred to the similar (top) and distant (bottom) German tasks (%
      \includegraphics[trim=35 17 439 334,clip,scale=0.441]{graph1.pdf} BASE,
      \includegraphics[trim=167 17 305 334,clip,scale=0.441]{graph1.pdf} PCA-NORM,
      \includegraphics[trim=352 17 120 334,clip,scale=0.441]{graph1.pdf} PCA-RAW%
      ).}
    \label{f:monolingual_results}
   \vspace{-3mm}
\end{figure}

Despite the fact that the LDA models were never trained on the target task data, the PCA-RAW and PCA-NORM representations yield high peaks in F1 when trained on 50 (F1 $68.9$, PCA-RAW) 
minimal pairs and tested on DE-ST (Figure \ref{f:monolingual_results}). PCA-RAW outperforms PCA-NORM for almost all data sizes. PCA-RAW outperforms the BERT (BASE) representations especially on the very low-resource setting (10--60 pairs), with an increase of F1 $+14.2$ at 40 pairs.
Once the training size reaches 70 pairs, the differences in F1 become smaller.
The subspace-based representations are especially useful for the low-resource scenario. 

\paragraph{DT: Distant Task}

For the distant task DT, the general F1 scores are lower than for the similar task ST. However, PCA-RAW still reaches a Macro-F1 of $63.5$ at 50 pairs
for DE-DT.
This indicates that the profane subspace found by PCA-RAW partially generalizes to a broader, offensive subspace. Similar to ST, the projected PCA-RAW representations are especially useful in the low-resource case up to 50 sentences. The F1 of the BERT baseline is well below the PCA-RAW representations when data is sparse, with a major gap of F1 $+10.9$
at 30 pairs for DE-DT.
The classifier using BASE representations stays 
around F1 $53.0$ (DE-DT)
and does not benefit from more data,
indicating that these representations do not generalize to the target tasks.
However, once normalization (PCA-NORM) is added, the generalization is also lost and we see a drop in performance around or below the baseline. As for ST, all three representation types level out once higher amounts of data (70--80 pairs) are reached.

The standard errors show a similar trend to those in the word-level experiments: we observe a small standard error when training data is sparse (10--40 pairs), indicating that the choice of minimal pairs has a small impact on the subspace quality, which decreases further 
when more minimal pairs are available for training (50--100 pairs).

\subsection{Zero-Shot Cross-Lingual Transfer}

\begin{figure*}
    \centering
    \includegraphics[width=\textwidth]{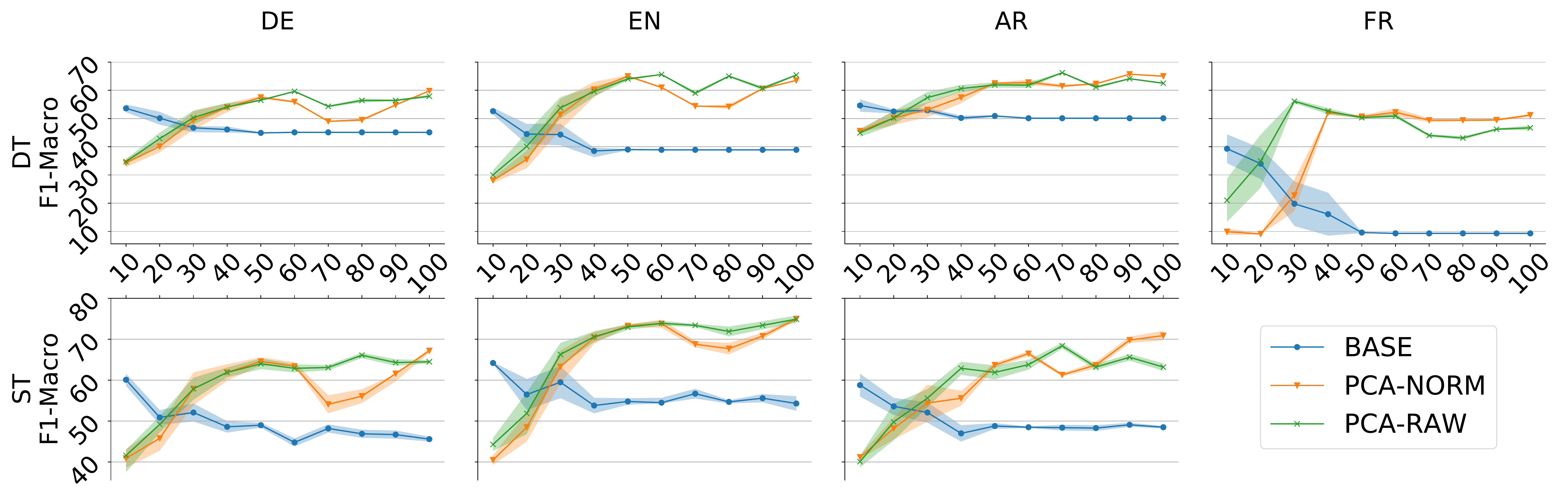}
    \caption{F1-Macro of the LDA model, using BASE or PCA-\{RAW,NORM\} representations, zero-shot transferred to the similar (bottom) and distant (top) German, English, Arabic and French tasks.}
    \label{f:multilingual_results}
\end{figure*}

To verify whether the subspaces also generalize to other languages, we zero-shot transfer and test the German BASE, PCA-RAW and PCA-NORM representations on the similar and distant tasks of closely-related (English), distantly-related (French) and non-related (Arabic) languages. For French, we only test on DT due to a lack of data for ST.

\subsubsection{Representation Types}

The setup is the same as in Section \ref{s:monolingual_representation}, except for
using \texttt{Bert-Base-Multilingual-Cased} and fine-tuning it on a corpus consisting of the 5M \{DE,EN,FR,AR\} tweets. 
The resulting
model is used to generate the hidden-representations needed to construct the BASE, PCA-RAW and PCA-NORM representations.
After performing 5-fold cross validation, the optimal number of PC is determined. Depending on the number of minimal pairs, the resulting subspace sizes lie between $8$--$67$ (PCA-RAW) and $10$--$44$ (PCA-NORM). 

\subsubsection{Results}

As in Section \ref{s:monolingual_results}, we train on increasingly large subsets of the German minimal pairs. 

\paragraph{ST: Similar Task}

We test the generalization of the German representations on the similar (\textit{neutral}/\textit{profane}) task on EN-ST and AR-ST as well as DE-ST for reference. Note that the LDA classifiers were trained on the German minimal pairs only, without access to target task data. 

The trends on the three test sets are very similar to each other (Figure \ref{f:multilingual_results}, bottom), indicating that the German profane subspaces transfer not only to the closely-related English, but also to the unrelated Arabic data. For all three languages, the PCA-\{RAW,NORM\} methods tend to grow in performance with increasing data until around 40 sentence pairs when the method seems to converge. This yields a performance of F1 $66.1$ on DE-ST at 80 pairs, F1 $74.9$ on EN-ST  at 100 pairs and F1 $68.4$
on AR-ST at 70 pairs for PCA-RAW. 

Overall, larger amounts of pairs are needed to reach top-performance in comparison to the monolingual case. This trend is also present when testing on DE-ST, leading us to posit that it is caused not by the cross-lingual transfer itself, but by the different underlying BERT models used to generate the initial representations. The differences in F1 between PCA-RAW and PCA-NORM are mere fluctuations between the two methods. The BASE representations are favorable only at 10 training pairs, with more data they overfit on the source task and are outperformed by the subspace representations, with differences of F1 $+20.6$ at 100 sentence pairs (PCA-RAW) on EN-ST, and F1 $+22.4$ at 100 sentence pairs (PCA-NORM) on AR-ST. 

\paragraph{DT: Distant Task}

Similar trends to ST are observed on the distant (\textit{neutral}/\textit{hate}) tasks (Figure \ref{f:multilingual_results}, top). While the BASE representations are strongest at 10 sentence pairs, they are outperformed by the subspace-based representations at around 30 pairs. PCA-RAW outperforms PCA-NORM and peaks at F1 $59.6$ (60 pairs), F1 $65.6$ (60 pairs), F1 $66.2$ (70 pairs) and F1 $56.1$ (30 pairs) for the German, English, Arabic and French test sets respectively.

We conclude that the German profane subspaces are transferable not only monolingually or to closely-related languages (English) but also to distantly-related (French) and non-related languages (Arabic), making a zero-shot transfer possible on both similar (neutral/profane) and distant tasks (neutral/hate). The BERT embeddings, on the other hand, were not able to perform the initial transfer, \ie from minimal-pair training to similar and distant target tasks, thus making the transfer to other languages futile. 
Subspace-based representations are a powerful tool to fill this gap, especially for classifiers trained on small amounts of source target data and zero-shot transfer to related tasks.

\paragraph{External Comparison}

The transfer capabilities of our subspace-based models can be set into perspective by comparing them to state-of-the-art classification models that were trained directly on our target tasks.
For \textbf{DT}, the top scoring team on EN-DT reaches higher levels of F1 ($75.6$) \citep{mandl2019hasoc} than our best PCA-RAW representations (F1 $65.6$). Similarly, the top scoring model on CHS-FR \citep{charitidis2019towards} lies at F1 $82.0$ and thus F1 $+25.9$ over PCA-RAW. However, PCA-RAW outperforms the best-performing model reported in \citet{mubarak-etal-2017-abusive} (F1 $60.0$) by F1 $+6.2$. Note, however, that this comparison is vague, as there is no standard train-test split for AR. For ST, no direct comparison to SOTA models can be made, since the profane-neutral classification task is usually part of a larger multi-class classification task. 
Nevertheless, the success of simple subspace-based LDA models, trained on very small amounts of task-distant German data, at cross-lingually zero-shot transferring to various tasks underlines the generalization capability of our approach.

\subsection{Qualitative Analysis}
A qualitative per-task analysis of the errors of the best performing models (PCA-RAW) reveals that some of the gold labels are debatable. The subjectivity of hate speech is a well-known issue for automatic detection tasks. Here, it is especially observable for EN, AR and FR, where arguably offensive comments were annotated as neutral but classified as offensive by our model:

\begin{quote}
   \textit{C'est toi la pute. Va voir ta mère \\
           \engl{You are the whore. Go see your mom}}
\end{quote}

We find that the models tend to over-blacklist tweets across languages as most errors stem from classifying neutrally-labeled tweets as offensive. This is triggered by negative words, \eg \textit{crime}, as well as words related to religion, race and politics, \eg:

\begin{quote}
    \textit{No Good Friday agreement, no deals with Trump.}
\end{quote}

\section{Conclusion and Future Work}
\label{s:conclusion}

In this work, we have shown that a complex feature such as \emph{profanity} can be encoded using semantic subspaces on the word and sentence-level. 

On the \textbf{word-level}, we found that the subspace-based representations are able to generalize to previously unseen words.
Using the profane subspace, we were able to substitute previously unseen profane words 
with neutral counterparts.

On the \textbf{sentence-level}, we have tested the generalization of our subspace-based representations (PCA-RAW, PCA-NORM) against raw BERT representations (BASE) in a zero-shot transfer setting on both similar (\textit{neutral}/\textit{profane}) and distant (\textit{neutral}/\textit{hate}) tasks.  While the BASE representations failed to zero-shot transfer to the target tasks, the subspace-based representations were able to perform the transfer to both similar and distant tasks, not only monolingually, but also to the closely-related (English), distantly-related (French) and non-related (Arabic) language tasks. 
We observe major improvements
between F1 $+10.9$ (PCA-RAW on DE-DT) and F1 $+42.9$ (PCA-NORM on FR-DT) over the BASE representations
in all scenarios.
As our experiments have shown that the commonly used mean-shift normalization
is not required, we plan to conduct further experiments using unaligned significant words/sentences.

The code, the fine-tuned models, and the list of minimal-pairs are made publicly available\footnote{\url{www.github.com/uds-lsv/profane_subspaces}}.

\section*{Acknowledgements}
We want to thank the annotators Susann Boy, Dominik Godt and Fabian G\"{o}ssl. We also thank Badr Abdullah, Michael Hedderich, Jyotsna Singh and the anonymous reviewers for their valuable feedback.
The project on which this paper is based was funded by the DFG under the funding
code WI 4204/3-1. Responsibility for the content of this publication is with the
authors.

\bibliographystyle{acl_natbib}
\bibliography{acl2021}

\end{document}